\documentclass[sigconf]{acmart}
\acmSubmissionID{850}

\usepackage{booktabs} 

\usepackage[ruled]{algorithm2e} 

\SetAlFnt{\small}
\SetAlCapFnt{\small}
\SetAlCapNameFnt{\small}
\SetAlCapHSkip{0pt}

\acmJournal{TOG}

\usepackage{makecell}

\newcommand{\Fig}[1] {Fig.\ #1}
\newcommand{\Figs}[1]{Figs.\ #1}
\newcommand{\Tbl}[1]  {Tab.\ #1}

\newcommand{\Sec}[1] {Sec.\ #1}
\newcommand{\SSec}[1] {Sec.\ #1}

\graphicspath{{./figures/}}

\copyrightyear{2023}
\acmYear{2023}
\setcopyright{acmlicensed}\acmConference[SA Conference Papers '23]{SIGGRAPH Asia 2023 Conference Papers}{December 12--15, 2023}{Sydney, NSW, Australia}
\acmBooktitle{SIGGRAPH Asia 2023 Conference Papers (SA Conference Papers '23), December 12--15, 2023, Sydney, NSW, Australia}
\acmPrice{15.00}
\acmDOI{10.1145/3610548.3618240}
\acmISBN{979-8-4007-0315-7/23/12}

\begin{document}
\title{360$^\circ$ Reconstruction From a Single Image Using Space Carved Outpainting}

\author{Nuri Ryu}
\orcid{0000-0002-7769-689X}
\affiliation{%
  \institution{POSTECH}
   \country{South Korea}
}
\email{ryunuri@postech.ac.kr}

\author{Minsu Gong}
\orcid{0009-0008-9304-5203}
\affiliation{%
  \institution{POSTECH}
   \country{South Korea}
}
\email{gongms@postech.ac.kr}

\author{Geonung Kim}
\orcid{0000-0003-0806-6963}
\affiliation{%
  \institution{POSTECH}
   \country{South Korea}
}
\email{k2woong92@postech.ac.kr}

\author{Joo-Haeng Lee}
\orcid{0000-0002-5788-712X}
\affiliation{%
  \institution{Pebblous}
   \country{South Korea}
}
\email{joohaeng@pebblous.ai}

\author{Sunghyun Cho}
\orcid{0000-0001-7627-3513}
\affiliation{%
  \institution{POSTECH}
   \country{South Korea}
}
\affiliation{%
  \institution{Pebblous}
   \country{South Korea}
}
\email{s.cho@postech.ac.kr}

\renewcommand{\shortauthors}{Ryu et al.}

\begin{teaserfigure}
    \centering
    \includegraphics[width=\linewidth]{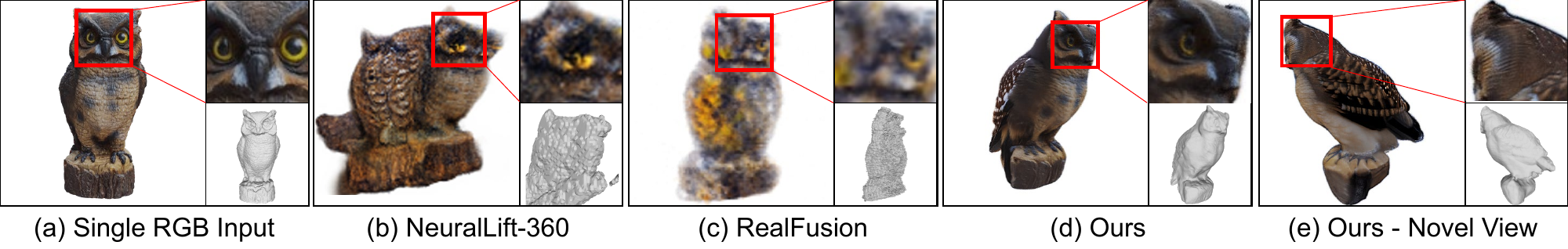}
    \vspace{-0.3cm}
    \caption{Examples of a full $360^\circ$ view 3D reconstruction from a single RGB image given in (a). The bottom-right image in (a) is the ground-truth mesh corresponding to the input image, while the other bottom-right images in (b)-(e) are reconstructed meshes by each method. 
    The results in (b) and (c) show that the na\"ive usage of the distillation loss and neural density fields leads to sub-optimal novel views and a low-fidelity surface~\cite{neuralLift,realfusion}. On the other hand, our framework successfully generates novel views that resemble the original input image and also reconstructs the 3D object's surface with high fidelity, as we observe in (d) and (e).
    Image in (a): rendered from the data in the Objaverse dataset~\cite{objaverse} [\copyright Horton, CC BY].
    } 
    \label{fig:teaser}
\end{teaserfigure}

\begin{abstract}
We introduce POP3D, a novel framework that creates a full $360^\circ$-view 3D model from a single image. POP3D resolves two prominent issues that limit the single-view reconstruction. Firstly, POP3D offers substantial generalizability to arbitrary categories, a trait that previous methods struggle to achieve. Secondly, POP3D further improves reconstruction fidelity and naturalness, a crucial aspect that concurrent works fall short of. Our approach marries the strengths of four primary components: (1) a monocular depth and normal predictor that serves to predict crucial geometric cues, (2) a space carving method capable of demarcating the potentially unseen portions of the target object, (3) a generative model pre-trained on a large-scale image dataset that can complete unseen regions of the target, and (4) a neural implicit surface reconstruction method tailored in reconstructing objects using RGB images along with monocular geometric cues. The combination of these components enables POP3D to readily generalize across various in-the-wild images and generate state-of-the-art reconstructions, outperforming similar works by a significant margin.
Project page: \url{http://cg.postech.ac.kr/research/POP3D}.
\end{abstract}

%
%



\maketitle


\section{Introduction}
\label{sec:introduction}

The ability to generate high-quality realistic 3D models from minimal input is an ongoing challenge for various applications in computer graphics, vision, virtual reality, and augmented reality.Despite the recent advances in the area of multi-view reconstruction through the differentiable rendering of neural representations~\cite{SRN, DVR, nerf}, such methods rely heavily on vast amounts of images paired with camera parameters. While this reliance may yield impressive results, it inhibits practicality and accessibility, particularly in scenarios where obtaining multiple views of an object is impractical or impossible.

In real-world scenarios, a user might only have a single view of an object. For instance, the image may be a photo of an object that is not easily accessible or an output of a 2D generative model. Consequently, 3D model generation from a single image has immense practical significance, enabling a broader range of applications and making 3D modeling more accessible to a wider user base.

Due to its practical significance, 3D reconstruction from a single image has been an active area of research. However, existing methods still suffer from two major problems: generalizability and reconstruction fidelity. Various methods have been proposed to learn from 3D data or object-centric videos for single-view reconstruction~\cite{GenerativePredictableVoxels, pix2vox, 3D-R2N2, AtlasNet, pixel2mesh, pifu, pointe, MCC}.
However, the acquisition of such data is often more challenging compared to collecting unstructured 2D data, thereby undermining the scalability and generalizability of these methods. While other techniques have also been proposed to circumvent the need for 3D data by relying on 2D image data, such methods are often bound to a specific category~\cite{codenerf, Shelf-sup, platonicgan, magicpony, nerf-from-image}, thereby limiting their generalizability. 

Concurrent methods \cite{realfusion, neuralLift, nerdi} that leverage a large-scale image prior \cite{StableDiffusion} via a distillation loss \cite{DreamFusion} frequently fall short of faithfully reconstructing the input view. This discrepancy arises as the distillation loss interferes with the RGB reconstruction loss of the input view and their limited target resolution of the reconstruction further exacerbates this problem. Furthermore, their use of na\"{i}ve neural density fields often leads to low-fidelity surface reconstruction.

This paper presents POP3D (Progressive OutPainting 3D), a novel framework designed to address the aforementioned issues of generalizability and reconstruction fidelity.To tackle the challenge of generalizability, our framework leverages the power of various priors pre-trained on large-scale datasets. This approach effectively mitigates the inherent ill-posedness of 3D reconstruction from a single RGB image across arbitrary categories.For a high-fidelity reconstruction covering the full $360^\circ$ view of an object, we generate novel views that match the quality of the given view through a large-scale generative model. These novel views, in conjunction with their monocular geometry predictions, form a pseudo-ground-truth dataset. By training on this dataset following a training strategy tailored to incorporate monocular geometry cues, we reconstruct a neural implicit surface and its corresponding appearance of the given single image with high fidelity compared to concurrent works as we illustrate in \Fig{~\ref{fig:teaser}}.

To elaborate, our framework begins by processing a single RGB input by using state-of-the-art monocular depth and normal predictors~\cite{ZoeDepth, Omnidata} to infer its geometric cues. The input RGB and its monocular geometry predictions constitute an initial dataset and are used to train an initial 3D model following a training strategy of MonoSDF~\cite{MonoSDF}. 
After initializing the 3D model, we update our camera position following a camera schedule that encompasses the full $360^\circ$ view of the target object. 
Then, our framework finds the visual hull~\cite{visual_hull} of the object, thereby computing the target object's seen area as well as the potentially unseen area. By removing the seen area, we obtain an outpainting mask, which is used as a guide for the generative model to produce a natural novel view of the object. Specifically, we use a conditional diffusion model~\cite{StableDiffusion} trained on a large-scale dataset~\cite{LAION-5B} capable of outpainting the image given a mask and a text condition. 
After a process of extracting the monocular geometry information of the outpainted result, we expand our pseudo-ground-truth dataset with the processed data.
This updated dataset is then used to retrain our 3D model and we repeat this gradual outpainting process until we create a dataset that covers a full $360^\circ$ view of an object, ultimately leading to a high-fidelity $360^\circ$ 3D reconstruction.

Our framework provides some distinctive benefits.
Firstly, thanks to the priors learned on large-scale datasets, our framework is not limited to a certain category of objects but can handle a wide range of objects from arbitrary categories.
Secondly, our framework does not need any additional external training data such as multi-view images or 3D geometries, as we adopt priors already learned in off-the-shelf models.
Thirdly, our progressive outpainting approach that builds a $360^\circ$-view dataset of the target object ensures the generation of novel-view images of high quality and the faithful reconstruction of the input image. 
Finally, by using the pseudo-ground-truth dataset to train a neural implicit surface representation, we can extract a well-defined high-quality surface.

To summarize, our primary contributions are as follows:
\begin{itemize}
	\item  We introduce a novel framework to reconstruct a full $360^\circ$ model from a single image. Our framework generalizes well to in-the-wild RGB images without any category-specific pre-training by leveraging off-the-shelf priors. 
	\item We develop a progressive outpainting scheme to generate pseudo-ground-truth images for 3D model reconstruction. Our method ensures a faithful reconstruction with novel-view images that naturally harmonize with the input image. Our model design accounts for both geometric and photometric consistency leading to high-fidelity shape and appearance reconstruction.
	\item We show that our framework can produce state-of-the-art $360^\circ$ reconstruction results from single RGB images in terms of novel-view synthesis and geometry reconstruction.
 \end{itemize}

\section{Related Works}

\subsection{Few-View-to-3D Reconstruction}

NeRF~\cite{nerf} and its variants~\cite{NeRF++, MipNeRF, MipNeRF360, RefNeRF} have shown remarkable reconstruction performance of scenes and objects only given RGB images paired with camera poses. However, without dense camera views, training a neural radiance field becomes a severely under-constrained problem. When only given a few views, such models may overfit to each given view resulting in a broken geometry and blurry noise when rendering novel views~\cite{DietNeRF}. Recently, a line of work has been introduced to reduce the number of required views for high-fidelity reconstruction~\cite{DietNeRF, RefNeRF, PixelNeRF, DDP, SRT, sparsefusion}. Nevertheless, they still require more than a single view for proper reconstruction.

\subsection{Single-View-to-3D Reconstruction}

Most of the early work that reconstruct 3D models from a single image rely on the visible information given in an image such as shading~\cite{sfs_a_survey}, texture~\cite{texture3d}, or defocus~\cite{blur3d}. Recent works use a more general prior in order to generate the invisible parts of an input image. For instance, some methods use 3D datasets to learn a 3D prior that can be used for reconstruction~\cite{GenerativePredictableVoxels, pix2vox, 3D-R2N2, AtlasNet, pixel2mesh, pifu}. However, a large-scale 3D dataset is needed for such models to generalize to in-the-wild images. Compared to large-scale 2D image datasets such as LAION-5B that offers 5.85 billion image-text pairs~\cite{LAION-5B}, 3D datasets are often limited in variety and scale. On the other hand, our model does not require any 3D training data but can generalize to in-the-wild images by leveraging geometry and image priors \cite{StableDiffusion,ZoeDepth, Omnidata} trained on large-scale datasets.

To overcome the issues arising from needing a 3D training dataset, methods that learn 3D structures from image collections have been introduced~\cite{CMR, Shelf-sup, platonicgan, magicpony, nerf-from-image, fast_and_explicit, vit_for_nerf, nerfdiff, holodiffusion, GeNVS, codenerf,ss3d}. However, they either need further annotations such as semantic key points and segmentation masks~\cite{CMR} or multi-view images of the same scene with accurate camera parameters~\cite{fast_and_explicit, vit_for_nerf, nerfdiff, holodiffusion, GeNVS,ss3d}. Other methods that train with single view per scene are category-specific~\cite{codenerf, Shelf-sup, platonicgan, magicpony, nerf-from-image}. In contrast, our model does not require any additional information apart from a single RGB image thanks to the off-the-shelf models that we incorporate. Also, we stress that our model can generalize to in-the-wild images regardless of the given view's category. 

While 3D diffusion models~\cite{TriplaneDiffusion, rodin} are also gaining attention, concurrent works~\cite{nerdi, neuralLift, realfusion, makeit3d} attempt to directly use a 2D diffusion model~\cite{StableDiffusion} trained on a large-scale image-text dataset~\cite{LAION-5B} as a prior for single view reconstruction. To generate unseen regions from the reference view, they heavily rely on a distillation loss similar to the score distillation sampling loss introduced by Poole et al.~\shortcite{DreamFusion}. The problem is that the distillation loss is simultaneously applied to views that have overlapping regions from the given single view. This often disrupts the RGB loss and consequently often leads to a poor reconstruction of the input view. While a very recent work \cite{makeit3d} tries to bypass this problem by projecting the reference image on to the trained 3D representation, novel views far from the reference view tend to lack quality. In contrary, our framework builds up a pseudo-ground-truth dataset composed of multi-view images that allow for the use of multi-view reconstruction strategies leading to high-fidelity reconstructions.

Furthermore, these models~\cite{nerdi, neuralLift, realfusion, makeit3d} have other limitations as well.
Firstly, they have a low target resolution, e.g.,  $96 \times 96$ or $128 \times 128$, while our method aims for a resolution of $384 \times 384$ yielding results with higher quality and overall detail. While Tang et al.~\shortcite{makeit3d} try to overcome the this problem through a two-stage training scheme, it shares the other problems described below as well.
Secondly, these works use naive neural density functions as their geometry representations, which may produce noisy artifacts due to the lack of a well-defined surface threshold. In contrast, our method simply allows for high-fidelity geometry extraction from the zero-level set of the learned neural implicit surface. Lastly, these models only rely on the given single image and its augmentations to personalize the diffusion model using a method similar to Textual Inversion~\cite{textual_inversion} in an attempt to generate unseen regions consistent with the input image. In contrast to these methods, our data generation framework allows the use of a state-of-the-art diffusion model personalization method, DreamBooth~\cite{dreambooth}, that requires multiple views of the same object by using multi-view pseudo-ground-truth images, which allows for a better personalization quality. 
Raj et al.~\shortcite{dreambooth3d} also showed that high-quality personalized text-to-3D can be achieved using DreamBooth. However, their method requires multiple views of a target object whereas our method only requires a single view of an object thanks to our pseudo-ground-truth multi-view generation scheme.

\paragraph{Single View to Point Cloud} Other recent works aim to reconstruct colorized point clouds based on a reference view. For instance, MCC~\cite{MCC} takes an RGB-D image as input and reconstructs the lifted color points into a complete point cloud of the target object. Similarly, Point-E~\cite{pointe} introduces a point-cloud diffusion model that uses a reference RGB image to generate a colorized point cloud that resembles the input image. Unlike such models, our framework reconstructs a high-fidelity neural implicit surface and an appearance of superior quality.

\paragraph{3D Photography} Another line of work utilizes monocular depth predictions and color inpainting to generate a 3D photo or scene from a single image~\cite{AdaMPI, 3DPhoto, text2room, text2nerf}.  
However, such methods are only designed to inpaint both the foreground and background at the same time, and does not account for the backside of an object. Therefore, they are not directly applicable to $360^\circ$ reconstruction of an object.

\paragraph{Novel View Synthesis from a Single View}
Some works~\cite{3DiM, zero1to3} focus on generating a 3D novel view when given an input image and a relative pose. However, the outputs of such models are only approximately 3D consistent and therefore do not guarantee a high-fidelity shape reconstruction. 

\begin{figure*}[t!]    
\begin{center}
\includegraphics[width=0.95\linewidth]{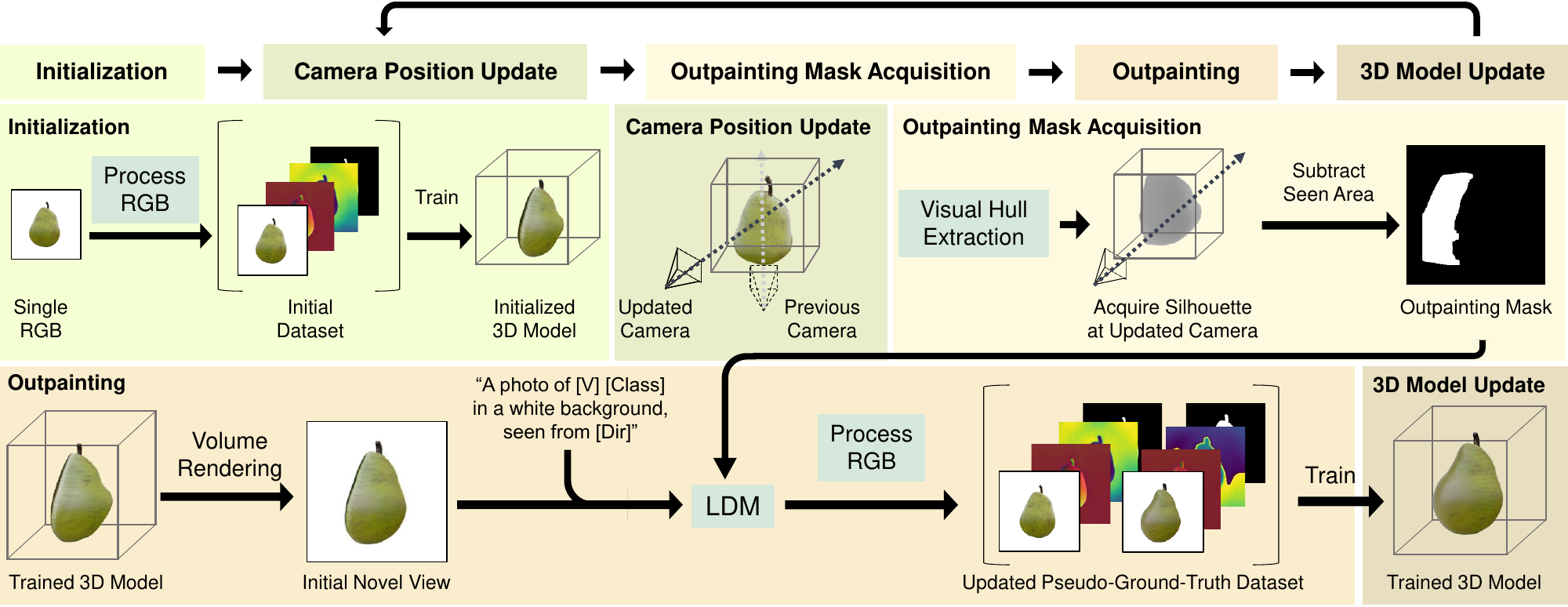}
\end{center}
\vspace{-4.0mm}
\caption{\textbf{Framework overview.} POP3D operates in five interconnected steps. Initially, we process the single RGB input to create a preliminary pseudo-ground-truth dataset and use this data to initialize a 3D model. We then progress through a loop of steps aiming to cover the complete $360^\circ$ view of the target object. This loop includes: updating the camera position according to a predetermined schedule; acquiring an outpainting mask by extracting the visual hull from the pseudo-ground-truth dataset and subtracting the seen area; generating a pseudo-ground-truth novel view using the initial novel view from the trained 3D model, outpainting mask, and a suitable text prompt; and training the 3D model using the updated pseudo-ground-truth dataset. This process continues until we encompass the $360^\circ$ view of the object.
Input image: rendered from the data in the Objaverse dataset~\cite{objaverse} [\copyright laboratorija, CC BY-NC-SA].
}
\label{fig:overview}
\end{figure*}
\begin{figure}[t!]    
\begin{center}
\includegraphics[width=1.0\linewidth]{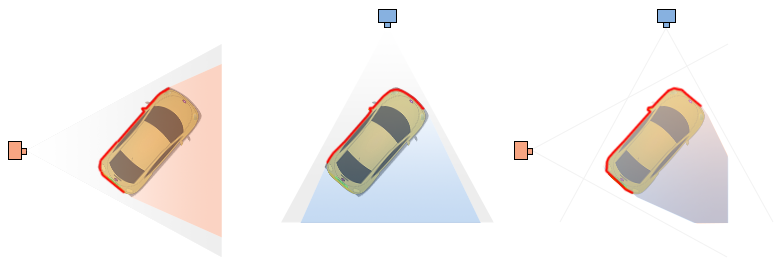}
\end{center}
\vspace{-6.0mm}
\caption{\textbf{Visual hull extraction}. We illustrate the acquisition of the visual hull from two camera views. We preserve the shaded region seen on the right which constitutes both seen and potentially unseen regions of a target object. Car image: from freesvg [Public Domain].}
\label{fig:visual_hull}
\end{figure}

\section{Method}

We present an overview of POP3D in Fig.~\ref{fig:overview}. Given a single image of an object, our framework reconstructs its 360$^\circ$ shape and appearance using a neural implicit surface representation.
Our key idea is to progressively outpaint the unseen regions of the object by synthesizing their color and geometric information.
To this end, our framework consists of five steps: initialization, camera position update, outpainting mask acquisition, outpainting, and 3D model update. In the initialization step, we estimate the depth and normal maps of the input image and lift it to a 3D view. Then, we update the camera position to a nearby viewpoint that has not been seen before, and obtain an outpainting mask that indicates the region to be outpainted using space carving~\cite{space_carving}. Next, we outpaint the masked region by generating its color and geometric information using a latent diffusion model (LDM) \cite{StableDiffusion}. Finally, we update the 3D model of the object using the outpainted information. We repeat these steps until we cover 
the entire 360$^\circ$ of the object.

To represent the shape and appearance of a 3D object, we adopt VolSDF~\cite{VolumeSDF}, which represents a 3D object using a pair of neural networks.
Specifically, to represent the geometry of an object, we use a neural network modeling a signed distance function (SDF) $f_\theta : x \mapsto s $, which maps a 3D point $x \in \mathbb{R}^3$ to its signed distance $s \in \mathbb{R}$ to the surface. To account for the appearance, we use another neural network that models a radiance function $L_\theta \left( \textbf{x}, \mathbf{\hat{n}}, \mathbf{\hat{z}} \right) $ where $\mathbf{\hat{n}}$ is the spatial gradient of the SDF at point $\textbf{x}$. $\mathbf{\hat{z}}$ is the global geometry feature vector same as in Yariv et al.~\shortcite{IDR}.
Unlike VolSDF, we do not give the viewing direction as input to $L_\theta$ and ignore view-dependent color changes as a single image does not provide view-dependent lighting information and conventional outpainting methods do not account for view dependency. 

As 3D model generation from a single image is an extremely ill-posed task, we impose a couple of assumptions to restrict the possible outcomes of the reconstruction results.
First, we assume that the target object lies within a cube, which has its center at the origin, and edges of length 2 aligned with the coordinate axes, and initialize the object as a unit sphere following Atzmon and Lipman~\shortcite{SAL}.
We also assume a virtual camera looking at the target 3D object during our 3D reconstruction process.
Specifically, we place the camera on a sphere of radius $3$ to point at the origin and parameterize its position using spherical coordinate angles.
The field of view (FoV) of the camera is set to $60^\circ$ assuming that the camera parameters of the input image are not given.

In the following, we describe each step of our framework in detail.

\subsection{Initialization}
\label{subsec:initialization}
The initialization step constructs an initial 3D model from an input image $L_0$ of a target object.
Specifically, given $L_0$,
we first extract the foreground object by estimating a binary mask $M_0$ using an off-the-shelf binary segmentation method~\cite{TRACER}.
We then estimate the depth map $D_0$ and the normal map $N_0$ for the foreground object using off-the-shelf monocular depth and normal estimators~\cite{ZoeDepth,Omnidata}.
Using the estimated depth and normal maps, and binary mask, we estimate an initial 3D model. 
Specifically, we first initialize the pseudo-ground-truth dataset $\mathcal{P}$ as $\mathcal{P}=\{(L_0,D_0,N_0,M_0,\phi_0)\}$ where $\phi_0$ is the initial camera position,
and train the implicit representation of the initial 3D model $(f_\theta, L_\theta)$ using $\mathcal{P}$. The initial camera position is set to $\phi_0=(90^\circ,0^\circ)$ where the first and second angles are the polar and azimuthal angles, respectively, assuming that the initial image contains the frontal side of the target object.
The pseudo-ground-truth dataset is iteratively updated in the following steps to progressively reconstruct the 3D model of a target object.
For training the implicit representation, we adopt the approach of MonoSDF~\cite{MonoSDF} with a slight modification to consider the mask $M_0$. Refer to the supplementary material for more details on the training.

\subsection{Camera Position Update}
\label{subsec:camera_pos_update}

After the model has been initialized,
we iteratively update the 3D model exploring the unseen regions of the target object by changing the camera position. To this end, we define a camera schedule $\mathcal{S} = \bigl[ \phi_0, \phi_1, \ldots, \phi_s \bigr] $, designed to cover the $360^\circ$ view of the target object,
and update the camera position at each iteration according to $\mathcal{S}$.
In theory, the camera schedule may be an arbitrary set provided that it encompasses the complete $360^\circ$ view. However, we found that an excessively small or large interval may detrimentally affect the output.
Hence, we use an interval of $45^\circ$ degrees in our experiments, and discuss the adverse impacts of an overly granular or coarse camera schedule in \SSec{\ref{subsub:camera_schedule}}. In the rest of the section, we will denote the camera positions that have been explored until the $i$-th camera position in $\mathcal{S}$ as $\mathcal{S}_{0:i}$ such that $0 \leq i \leq s$. 

\subsection{Outpainting Mask Acquisition}
\label{subsec:out_mask_acq}

In order to generate the appearance and shape of unseen regions seamlessly, the areas designated for outpainting need to be appropriately chosen. To address this, we leverage the concept of the visual hull~\cite{visual_hull}. The visual hull provides a rough approximation of the object's shape derived from the object's silhouettes from different viewpoints. Using our current dataset $\mathcal{P}$, we can compute the visual hull to determine the maximum possible area that the object might occupy as illustrated in \Fig{\ref{fig:visual_hull}}. By computing the silhouette of the visual hull seen from the updated camera view, we obtain an initial mask that comprises both previously observed and potentially unseen regions. To create our outpainting mask, we subtract the observed regions from this initial mask, leaving only the potentially new visible areas.

\paragraph{Visual Hull Computation via Space Carving}
For the computation of the visual hull~\cite{visual_hull}, we use a depth-based voxel carving method driven by a voting scheme \cite{space_carving}. The process first voxelizes the object-bounding cube. Now we assume that we have explored the camera positions of $\mathcal{S}_{0:i-1}$.
Then, for a voxel $\mathbf{p}$ from the voxelized bounding cube, we raise a vote if its projection to the $j$-th view is inside the foreground region where $j \in \left\{0, \ldots, i -1 \right\}$, and if its distance to the $j$-th view's camera center is longer than the distance between the foreground region and the $j$-th view's camera center.
Mathematically, we raise a vote for $\mathbf{p}$ if
\begin{equation}
	\begin{aligned}
		K P_j \mathbf{p} &\in \hat{M}_j, \qquad\qquad\textrm{and} \\
		\| \mathbf{p} - \mathbf{o}_j \|_2 &\geq \| \mathbf{p^*} - \mathbf{o}_j \|_2, \\
	\end{aligned}
\end{equation}
where $K$ is the camera intrinsic matrix. $P_j$ and $\mathbf{o}_j$ represent the projection to the camera space and the camera center of the $j$-th view in $\mathcal{P}$, respectively. 
$\mathbf{p^*}$ is the point of intersection between the zero-level set of $f_\theta$ and the ray cast from $\mathbf{o}_j$ towards $\mathbf{p}$. Here, we only consider the first intersection where the ray penetrates the object from the exterior for the first time.
If the total number of vote counts equals the size of $\mathcal{P}$, or the number of views, the voxel is preserved. Otherwise, the voxel is carved out. Through this procedure, we collect the voxels comprising the visual hull of $\mathcal{P}$. To add, when we only have a single image, this process can be thought of an extrusion of the trained 3D surface. By projecting the visual hull onto the $i$-th viewpoint, we obtain its silhouette $M^{\text{VH}}_{i}$.

\paragraph{Foreground Mask Computation via Warping Operation}
Since $M^{\text{VH}}_i$ contains both seen and unseen regions, we should subtract out the seen region in order to obtain our outpainting mask $\tilde{M}_i$.
This is achieved by using a warping operation to compute the foreground mask $M^{\text{FG}}_{i}$ in the target view. The process involves rendering the depth from the previous viewpoints $\mathcal{S}_{0:i-1}$, lifting the image points to the 3D space, and subsequently projecting the lifted points to the target view $\phi_{i}$. To mitigate aliasing during the warping process, we scale up the image by a scaling factor of 8. We account for visibility and do not warp pixels not visible from the target viewpoint via back-point culling.

Specifically, the warping process including the back-point culling is performed as follows.
In the process of warping an image from the $j$-th view to the $i$-th view, we denote a lifted pixel from the $j$-th image in $\mathcal{P}$ as $\mathbf{p}^{j}$.
Then, we render its normal $\hat{N}^{j}$ from $f_\theta$. The pixel is only warped for the target camera center $\mathbf{o}_{}$ if:
\begin{equation}\label{eq:back-point culling}
	\left( \mathbf{p}^{j} - \mathbf{o}_{i} \right) \cdot \hat{N}^{j} < 0
\end{equation}
The warped coordinate $\textbf{p}^{i}$ is then computed as:
\begin{equation}
	\textbf{p}^{i} = K P_{j \to i}  K^{-1} \textbf{p}^{j}
\end{equation}
where $P_{j \to i}$ denotes a $4\times4$ transformation matrix that warps the camera from the source position to the target position. 
A binary mask $M^{\text{FG}}_{i}$ is computed from the collection of the warped pixels. Then, $\tilde{M}_{i}$ is calculated as $\tilde{M}_{i} = M^{\text{VH}}_{i} - M^{\text{FG}}_{i}$.

\subsection{Pseudo-Ground-Truth Generation}
\label{subsec:psg_gen}
In order to reconstruct the $360^\circ$ shape and appearance of the target object, we generate pseudo-ground-truth images to fill in the unseen parts of the object. For this purpose, we use a pretrained state-of-the-art generative model. Specifically, we use the Latent Diffusion Model (LDM) \cite{StableDiffusion} that takes an RGB image, a mask condition, and a text condition as input and outputs an RGB image following the input conditions.

However, na\"ively using a pretrained diffusion model may result in outpainting results that do not resemble the reference image. To generate pseudo-ground-truth images that are coherent to the given single view, we adopt a personalization technique outlined in DreamBooth \cite{dreambooth}. This technique allows us to learn a unique identifier token [V] of the object, which can be included as part of the text condition.
Since our framework generates multi-view pseudo-ground-truth images, we can trivially apply such a technique to generate well-harmonized results. For the details of the conditional diffusion model and its personalization, we refer the readers to the supplementary document. 

As the inputs to the personalized LDM, we use:
\begin{itemize}
	\item $I_i$ - the RGB image rendered from the trained model at the updated camera view,
	\item $\tilde{M}_i$ - the outpainting mask at the updated camera view, as detailed in Section \ref{subsec:out_mask_acq}, and
	\item a text prompt designed to generate view-consistent results.
\end{itemize}
For the text condition, we utilize a prompt structured as ``A photo of [V] [Class] in a white background, seen from [Dir]'' where [V] represents the personalized unique identifier of the specific object, [Class] refers to a simple class keyword such as `hamburger' or `doll', and [Dir] is a directional keyword such as `front', `left', `right' and `behind' used to guide the generation following the approach of Poole et al.~\shortcite{DreamFusion}.

Upon obtaining the outpainted view, we apply $2\times$ super-resolution \cite{DiffSR} for image enhancement. We then employ off-the-shelf monocular depth \cite{ZoeDepth} and normal \cite{Omnidata} estimators to extract geometric predictions $D_i$ and $N_i$. Furthermore, we use a background remover \cite{TRACER} to obtain the foreground mask $M_i$.
Finally, we update the pseudo-ground-truth dataset $\mathcal{P}$ as $\mathcal{P}\leftarrow \mathcal{P}\cup\{(L_i,D_i,N_i,M_i,\phi_i)\}$.

\subsection{3D Model Update}
Using the updated pseudo-ground-truth dataset $\mathcal{P}$, we train the SDF $f_\theta$ and neural radiance field $L_\theta$ following MonoSDF~\cite{MonoSDF}. After retraining the target 3D model, we return to the camera position update step described in \Sec{~\ref{subsec:camera_pos_update}}, and continue the loop until we go through the whole camera schedule $\mathcal{S}$.
\begin{figure}[t!]    
\begin{center}
\includegraphics[width=0.95\linewidth]{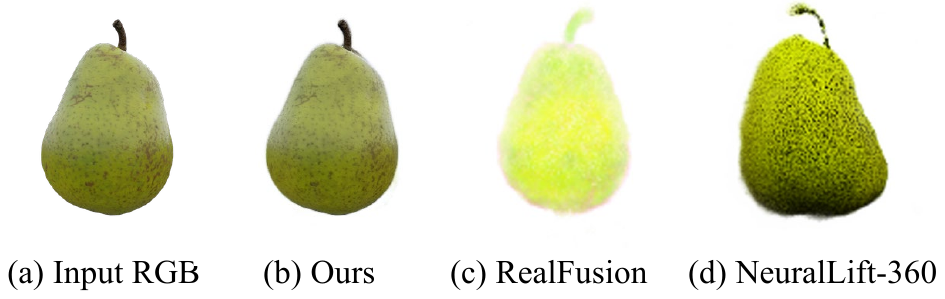}
\end{center}
\vspace{-3.0mm}
\caption{\textbf{Qualitative comparison on the input image reconstruction.} Given a single input image (a), our method successfully reconstructs the reference view as seen in (b). However, RealFusion~\cite{realfusion} and NeuralLift-360~\cite{neuralLift} do not faithfully reconstruct the input view even when it utilizes an RGB reconstruction loss.
Input image in (a): rendered from the data in the Objaverse dataset~\cite{objaverse} [\copyright laboratorija, CC BY-NC-SA].
} 
\label{fig:qual_recon}
\vspace{-4.0mm}
\end{figure}
\begin{figure}[t!]    
\begin{center}
\includegraphics[width=\linewidth]{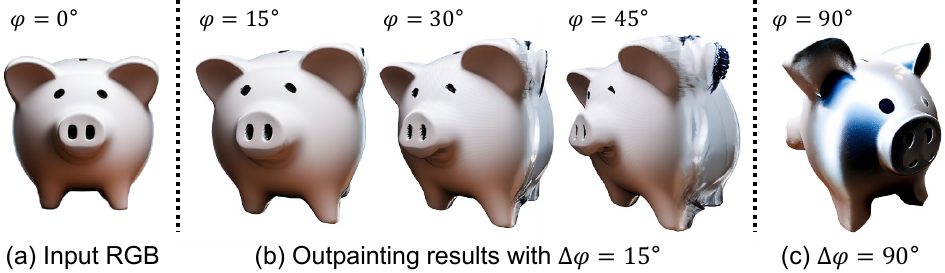}
\end{center}
\vspace{-3.0mm}
 \caption{\textbf{Effect of camera intervals on outpainting results}. For a single RGB input (a), both camera intervals excessively small (b) and overly large (c) have their drawbacks in their own ways as described in \Sec{~\ref{subsub:camera_schedule}}. Input RGB: generated using a diffusion model~\cite{StableDiffusion}.}
\label{fig:ablation_camera}
\end{figure}
\begin{figure}[t!]    
\begin{center}
\includegraphics[width=1.0\linewidth]{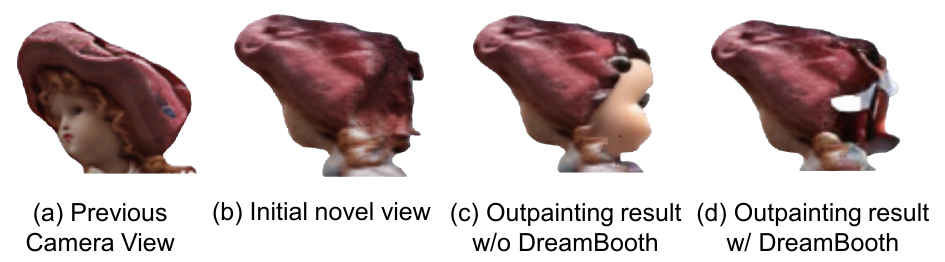}
\end{center}
\vspace{-3.0mm}
 \caption{\textbf{Effect of diffusion model personalization}. After updating the camera view from (a) to (b), we outpaint the target object with the same input to the LDM. However, na\"ive usage of the LDM may result in an outpainting result that does not consider the images in the pseudo-ground-truth dataset such as a doll with a duplicate face in (c), rather than naturally outpainting the doll's hat as shown in (d).
 Input image of this experiment: rendered from the data in the Objaverse dataset~\cite{objaverse} [\copyright shirava, CC BY].
 }
\vspace{-4.0mm}
\label{fig:ablation_dreambooth}
\end{figure}

\section{Experiments}

This section presents experimental results to evaluate the reconstruction quality of our framework in terms of shape and appearance. For all the experiments, we employ a camera schedule $\mathcal{S}$ whose polar angles are $90^\circ$ and the azimuthal angles are $\bigl[0^\circ, 45^\circ, $ $ -45^\circ, 90^\circ, -90^\circ, 135^\circ, -135^\circ, 180^\circ\bigr]$.
Nevertheless, our framework is not limited to a particular camera schedule. We show more qualitative results using a customized camera trajectory in the supplementary document.
For comparison, we utilize objects reconstructed using photogrammetry from Objaverse~\cite{objaverse}. Our selection of ten object categories provides a diverse range of shapes and appearances for testing.
We compare our method with concurrent works \cite{neuralLift, realfusion, MCC, pointe} that can reconstruct $360^\circ$ appearance and shape from a single reference view, along with a single-image-based 3D shape reconstruction method \cite{ss3d}.
To ensure a fair comparison, we use the same off-the-shelf depth estimator~\cite{ZoeDepth} for methods that require monocular depth guidance \cite{neuralLift, realfusion, MCC}.

\subsection{Comparisons with Other Methods}

\paragraph{Input-View Reconstruction}
Given a single RGB input, we expect the model to faithfully reconstruct the given image after the training process. Therefore, we compare our framework with methods that also use an RGB reconstruction loss during training to test the input-view reconstruction capability. 
To inspect the fidelity of the reference-view reconstruction, we use commonly used image quality metrics: PSNR, SSIM \cite{SSIM}, and LPIPS \cite{LPIPS}. Our method outperforms the others in all categories as shown in \Tbl{\ref{table:recon_cmp}}. As discussed in \Sec{~\ref{sec:introduction}}, concurrent works exhibit lower input-view reconstruction performance since the RGB loss is affected by the distillation loss in similar viewpoints. However, our framework that trains directly on a multi-view pseudo-ground-truth dataset does not face such a problem as we observe in \Fig{~\ref{fig:qual_recon}}.

\paragraph{Novel-View Synthesis}
For the evaluation of novel-view synthesis, we evaluate the results in terms of the similarity to the ground-truth views and the overall image quality of the output. CLIP similarity~\cite{CLIP} is used to evaluate the similarity between the model outputs and their corresponding ground-truth views. The image qualities of the generated outputs are assessed via the NIQE score~\cite{NIQE}. We evenly sample views around a $360^\circ$ trajectory, resulting in a total of 100 views for comparison. 

\begin{figure*}[t!]    
\begin{center}
\includegraphics[width=1.0\linewidth]{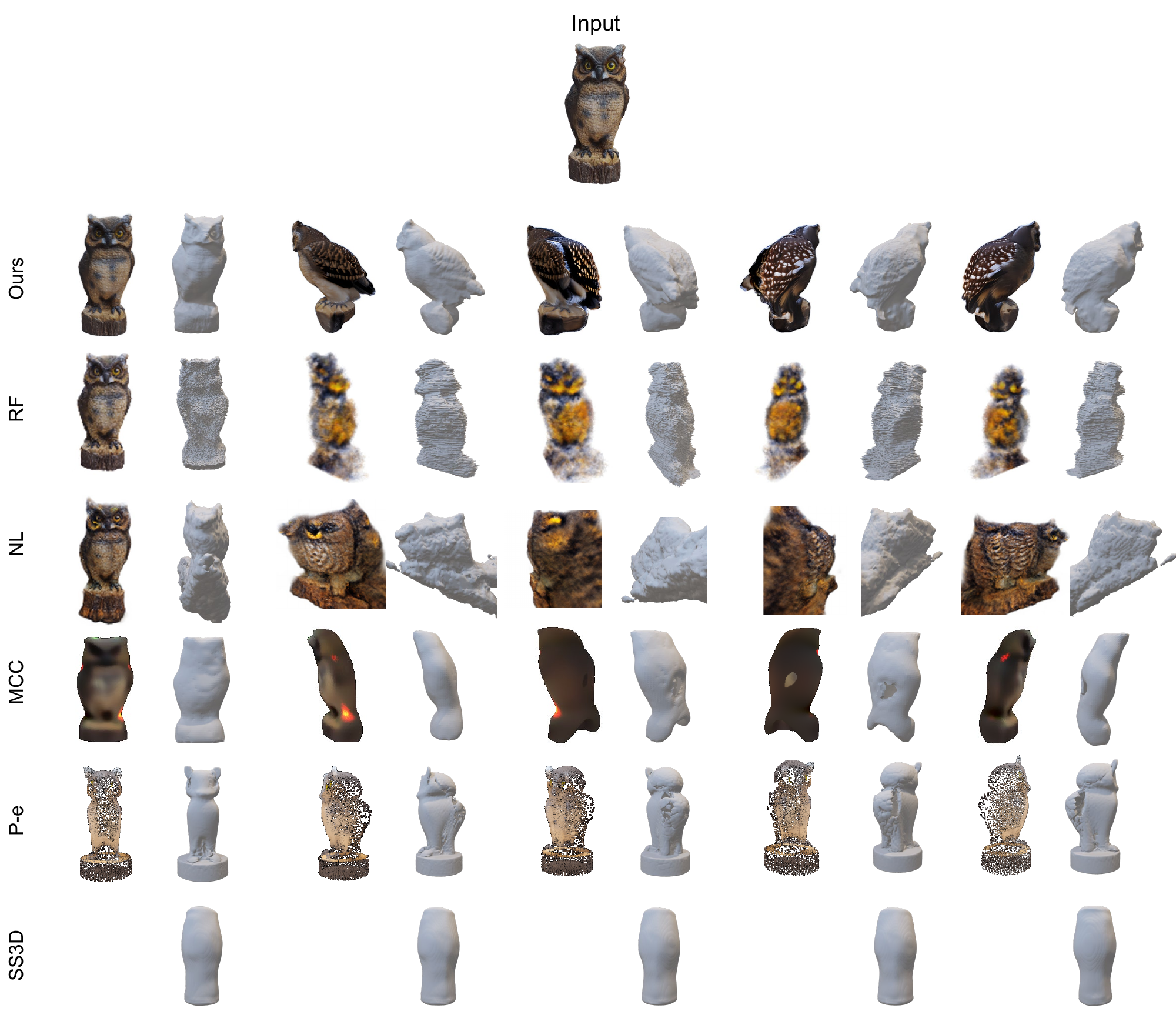}
\end{center}
\vspace{-4.0mm}
\caption{\textbf{Qualitative comparison.} We reconstruct the $360^\circ$ shape and appearance of the single RGB image given on top with various models and compare them with our result. NL, RF, MCC, P-e, and SS3D stand for NeuralLift~\cite{neuralLift}, RealFusion~\cite{realfusion}, MCC~\cite{MCC}, Point-E~\cite{pointe}, and SS3D~\cite{ss3d}, respectively. Since SS3D~\cite{ss3d} does not reconstruct the object's appearance we only show its shape output. For better visualization, we use marching cubes for MCC~\cite{MCC} to extract the surface with the same occupancy threshold that is used to sample the point cloud. For Point-E~\cite{pointe}, we use the point-cloud-to-mesh conversion provided by the authors. Input: rendered from the data in the Objaverse dataset~\cite{objaverse} [\copyright Horton, CC BY].}
\label{fig:qual_com_1}
\vspace{-4.0mm}
\end{figure*}

\begin{figure*}[t!]    
\begin{center}
\includegraphics[width=1.0\linewidth]{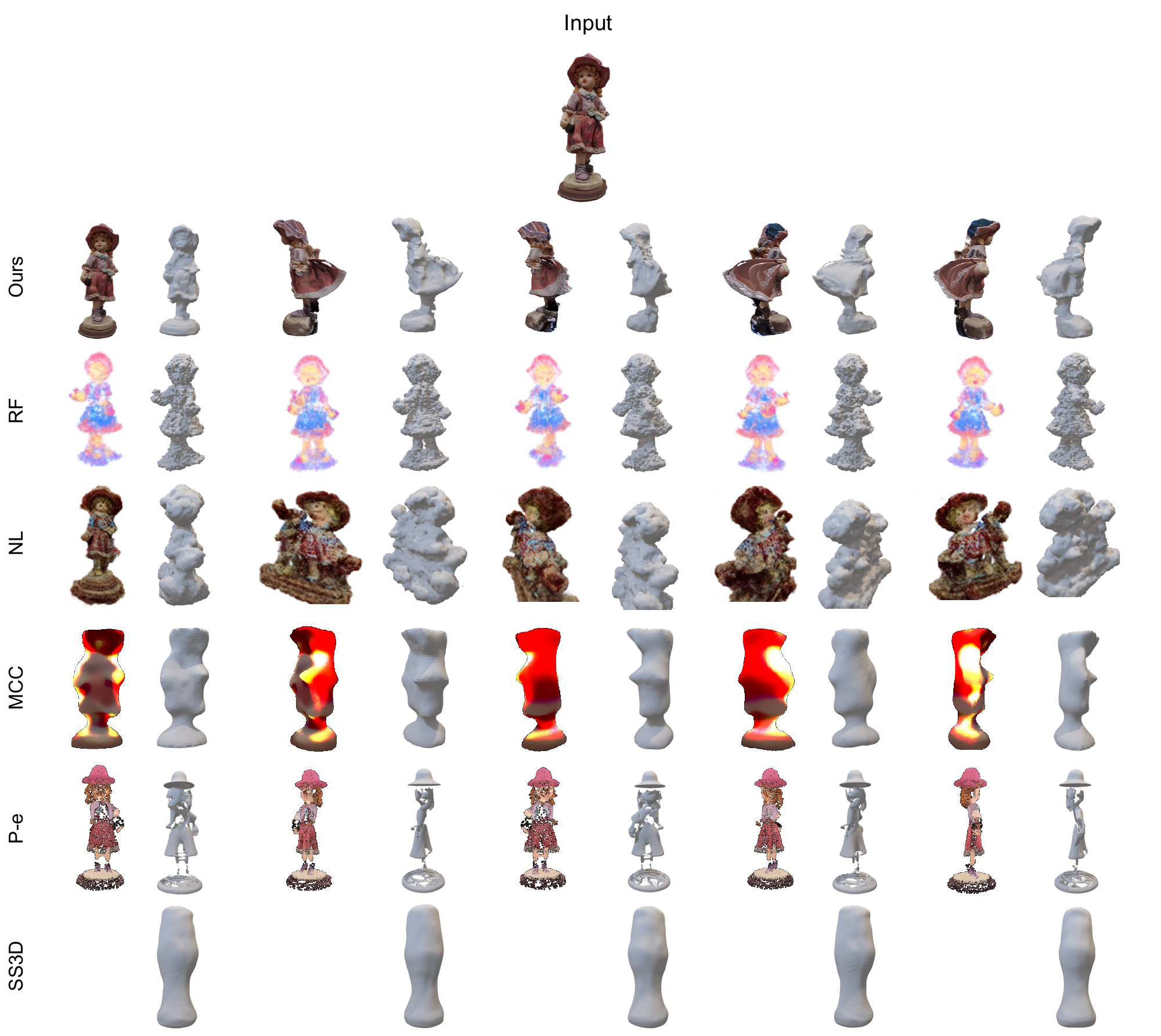}
\end{center}
\vspace{-4.0mm}
\caption{\textbf{Another qualitative comparison.} We reconstruct the $360^\circ$ shape and appearance of the single RGB image given on top with various models and compare them with our result. NL, RF, MCC, P-e, and SS3D stand for NeuralLift~\cite{neuralLift}, RealFusion~\cite{realfusion}, MCC~\cite{MCC}, Point-E~\cite{pointe}, and SS3D~\cite{ss3d}, respectively. Since SS3D~\cite{ss3d} does not reconstruct the object's appearance we only show its shape output. For better visualization, we use marching cubes for MCC~\cite{MCC} to extract the surface with the same occupancy threshold that is used to sample the point cloud. For Point-E~\cite{pointe}, we use the point-cloud-to-mesh conversion provided by the authors. Input: rendered from the data in the Objaverse dataset~\cite{objaverse} [\copyright shirava, CC BY].}
\label{fig:qual_com_doll}
\vspace{-4.0mm}
\end{figure*}

\Tbl{\ref{table:nvs_cmp}} presents a quantitative comparison.
As shown in the table, our method consistently shows high CLIP scores and outperforms the others in NIQE scores by a large margin for all categories. 
This shows that our method can generate novel views that are semantically similar to the given single view while maintaining high quality.
Qualitative comparisons of novel-view synthesis and their corresponding shapes are presented in \Figs{~\ref{fig:qual_com_1}} and \ref{fig:qual_com_doll}, where it can be observed that our method generates natural-looking novel views throughout the entire $360^\circ$ trajectory. In contrast, concurrent methods \cite{neuralLift, realfusion} often produce results that hardly resemble the input RGB images since their diffusion model personalization only relies on the single input view and its augmentations. In contrast, we leverage diffusion model personalization using multiple generated views, leading to a more coherent output. Moreover, our framework's utilization of neural implicit surface representation effectively reduces the introduction of foggy artifacts commonly seen in methods that adopt a more simplistic use of neural density fields. Compared to the methods that reconstruct colorized point clouds \cite{MCC, pointe}, our framework generates novel views with much finer details. 

\subsection{Ablations}

\subsubsection{Outpainting Errors and the Camera Schedule Interval}
\label{subsub:camera_schedule}

While we may use any camera schedule as long as it covers the entire 360$^\circ$ of a target object in theory,
during the outpainting process, we identified two key factors that may precipitate failure scenarios. 
The first issue arises when the outpainting mask barely extends beyond the object's boundary. In this instance, the input image dominates the input condition, making the outpainting process highly sensitive to any artifacts in the immediate vicinity of the outpainting region. By repeating the outpainting process, such boundary artifacts are accumulated, which often leads to failure cases.
The second issue manifests when the outpainting mask is excessively large compared to the object region in the original image. In this case, the outpainting tends to generate an image that adheres largely to the text prompt, thus neglecting the input image.

Consequently, excessively granular or large camera intervals may result in reconstruction failures as depicted in \Fig{~\ref{fig:ablation_camera}}. To mitigate these issues, we use an interval size of $45^\circ$ in our experiments. Empirically, this interval size effectively circumvents outpainting failures, thereby facilitating the reconstruction of high-fidelity $360^\circ$ views of a target object. Nevertheless, while the suggested interval size may serve as a good starting point, the best warping angle interval or the camera schedule itself may vary for various objects. We show examples of more customized intervals in the supplementary.

\subsubsection{Outpainting Results Without LDM Personalization}
\label{subsub:dreambooth_ablation}
As our model architecture generates multiple pseudo-ground-truth views of a target object throughout the reconstruction process, it allows for the personalization of the pre-trained LDM \cite{StableDiffusion} using the state-of-the-art technique, DreamBooth~\cite{dreambooth}. 
The benefit of this approach is evidenced in \Fig{~\ref{fig:ablation_dreambooth}}, where the application of the personalized LDM \cite{StableDiffusion} generates a natural-looking novel view that seamlessly integrates with the pseudo-ground-truth dataset. In contrast, na\"ive reliance on the vanilla LDM may result in an outpainting outcome that does not reflect the previously seen views of the target object.

\begin{table}[t!]
\centering
\caption{\textbf{Quantitative comparison of the PSNR, SSIM and LPIPS scores}. Our model shows a large margin in terms quantitative result of the reference view reconstruction. NL and RF stand for NeuralLift~\cite{neuralLift} and RealFusion~\cite{realfusion}, respectively.}
\vspace{-0.3cm}
\resizebox{\columnwidth}{!}{
\begin{tabular}{c|ccc|ccc|ccc}

\Xhline{2\arrayrulewidth}  & \multicolumn{3}{c|}{PSNR $\uparrow$ } & \multicolumn{3}{c|}{SSIM $\uparrow$ } & \multicolumn{3}{c}{LPIPS $\downarrow$ } \\
                & NL     & RF     & Ours            & NL    & RF    & Ours           & NL    & RF    & Ours                    \\ \hline \hline 
Berry           & 17.47  & 27.85  & \textbf{32.30}  & 0.88  & 0.94  & \textbf{0.99}  & 0.21    & 0.12    & \textbf{0.02}      \\
Broccoli        & 18.28  & 14.73  & \textbf{37.66}  & 0.85  & 0.83  & \textbf{0.99}  & 0.24    & 0.28    & \textbf{0.01}      \\
Cactus          & 17.33  & 22.99  & \textbf{31.50}  & 0.90  & 0.92  & \textbf{0.98}  & 0.17    & 0.16    & \textbf{0.02}      \\
Cauliflower     & 15.50  & 27.81  & \textbf{33.60}  & 0.86  & 0.93  & \textbf{0.99}  & 0.23    & 0.16    & \textbf{0.02}      \\
Croissant       & 12.60  & 29.68  & \textbf{36.60}  & 0.82  & 0.96  & \textbf{0.99}  & 0.29    & 0.10    & \textbf{0.01}      \\
Doll Statue     & 14.68  & 13.89  & \textbf{39.49}  & 0.85  & 0.87  & \textbf{0.99}  & 0.21    & 0.23    & \textbf{0.01}      \\
Frog Statue     & 14.62  & 20.27  & \textbf{36.70}  & 0.90  & 0.91  & \textbf{0.99}  & 0.20    & 0.19    & \textbf{0.01}      \\
Owl             & 16.41  & 27.75  & \textbf{36.01}  & 0.84  & 0.91  & \textbf{0.99}  & 0.23    & 0.17    & \textbf{0.01}      \\
Pear            & 10.26  & 15.86  & \textbf{40.85}  & 0.67  & 0.88  & \textbf{0.99}  & 0.44    & 0.17    & \textbf{0.01}      \\
Skull           & 1.79   & 24.99  & \textbf{36.30}  & 0.14  & 0.94  & \textbf{0.99}  & 0.80    & 0.13    & \textbf{0.02}      \\ \hline
Mean            & 13.89  & 22.58  & \textbf{36.10}  & 0.77  & 0.91  & \textbf{0.99}  & 0.30    & 0.17    & \textbf{0.01}      \\
 \Xhline{2\arrayrulewidth}
\end{tabular}
}
\label{table:recon_cmp}
\end{table}

\begin{table}[t!]
\centering
\caption{\textbf{Quantitative comparison of the CLIP similarity and NIQE scores}. Our model not only achieves the best embedding similarity but also achieves the best image quality score. NL, RF, MCC, and P-e stand for NeuralLift~\cite{neuralLift}, RealFusion~\cite{realfusion}, MCC~\cite{MCC}, and Point-E~\cite{pointe}, respectively.}
\vspace{-0.3cm}
\resizebox{\columnwidth}{!}{
\begin{tabular}{c|ccccc|ccccc}

\Xhline{2\arrayrulewidth}  & \multicolumn{5}{c|}{CLIP $\uparrow$ } & \multicolumn{5}{c}{NIQE $\downarrow$ }\\
             & NL   & RF   & MCC  & P-e  & Ours             & NL   & RF   & MCC  &P-e   & Ours  \\ \hline \hline 
Berry        & 0.77 & 0.81 & 0.76 & \textbf{0.84} & 0.82             & 22.81 & 9.56 & 23.35 & 19.33 & \textbf{5.80}  \\
Broccoli     & 0.86 & \textbf{0.88} & 0.69 & 0.80 & 0.84             & 27.79 & 11.65 & 20.28 & 21.57 & \textbf{9.04}  \\
Cactus       & 0.79 & 0.82 & 0.69 & 0.84 & \textbf{0.89}             & 24.92 & 15.39 & 26.57 & 22.70 & \textbf{9.36}  \\
Cauliflower  & 0.82 & 0.80 & 0.74 & 0.77 & \textbf{0.87}             & 22.58 & 12.76 & 22.61 & 19.91 & \textbf{7.37}  \\
Croissant    & 0.74 & 0.76 & 0.70 & 0.78 & \textbf{0.86}             & 29.07 & 13.86 & 24.10 & 32.01 & \textbf{12.28}  \\
Doll Statue  & 0.78 & 0.77 & 0.67 & \textbf{0.85} & 0.84    & 21.46 & 21.75 & 24.49 & 31.42 & \textbf{18.79}  \\
Frog Statue  & 0.76 & 0.77 & 0.81 & 0.83 & \textbf{0.85}    & 22.30 & 15.13 & 18.22 & 19.63 & \textbf{8.77}  \\
Owl          & 0.81 & 0.76 & 0.67 & 0.80 & \textbf{0.86}             & 11.19 & 11.82 & 22.99 & 16.93 & \textbf{7.16}  \\
Pear         & 0.81 & 0.86 & 0.81 & 0.82 & \textbf{0.88}             & 26.36 & 10.49 & 25.01 & 17.88 & \textbf{9.50}  \\
Skull        & 0.71 & 0.83 & 0.76 & 0.81 & \textbf{0.87}             & 24.62 & 13.25 & 19.85 & 16.18 & \textbf{7.74}  \\\hline
Mean         & 0.78 & 0.80 & 0.73 & 0.81 & \textbf{0.86}             & 23.31 & 13.57 & 22.75 & 21.76 & \textbf{10.00}  \\
 \Xhline{2\arrayrulewidth}
\end{tabular}
}
\label{table:nvs_cmp}
\vspace{-4mm}
\end{table}

\begin{figure}[t!]    
\begin{center}
\includegraphics[width=0.9\linewidth]{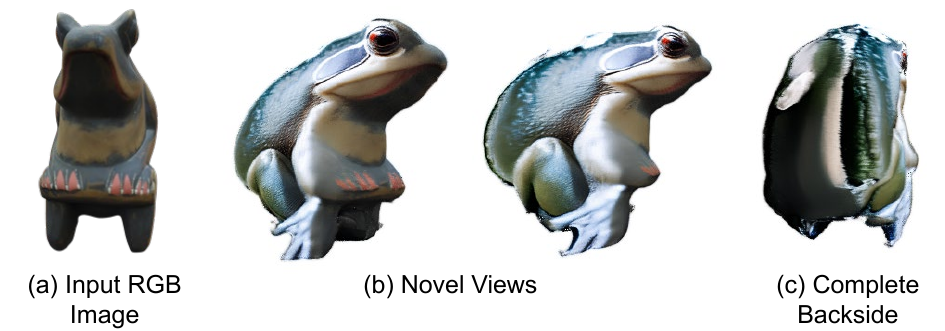}
\end{center}
\vspace{-4.0mm}
 \caption{\textbf{Limitations}. Given an input RGB (a) on the left, our model generates plausible novel views (b) following a camera schedule. However, our model sometimes generates a subpar complete backside image when compared to the input single RGB, as we see in (c).
  Image in (a): rendered from the data in the Objaverse dataset~\cite{objaverse} [\copyright ShekhirevaVictoria, CC BY].}
\vspace{-4.0mm}
\label{fig:limitation}
\end{figure}

\section{Conclusion and Future Work}

In this study, we present POP3D, a novel framework that addresses two long-standing challenges in the domain of $360^\circ$ reconstruction from a single RGB image: generalization and fidelity. POP3D fully leverages current state-of-the-art priors trained on large-scale datasets and successfully overcomes the problem of generalization to arbitrary images. 
Experimentally, we demonstrate that our framework not only faithfully reconstructs the provided single RGB input but also generates realistic novel views. These views collectively form a pseudo-ground-truth multi-view dataset, facilitating the direct application of multi-view reconstruction strategies. Compared to existing methodologies, our approach exhibits superior performance, reaffirming its potential as a robust solution for 3D reconstruction tasks.

\subsection{Limitations}

Our approach does exhibit certain limitations. Since our framework is a composition of off-the-shelf priors each playing a significant role in the pipeline, a failure of one model may impact the final reconstruction result.
For instance, if the monocular depth or normal predictors fail on challenging cases, e.g., thin structures, this may lead to artifacts in the reconstructed shape.

Moreover, our framework occasionally generates the object's complete backside with subpar quality when compared to the input view. We illustrate this problem in \Fig{~\ref{fig:limitation}}. This deficiency may be attributed to an accumulation of outpainting artifacts, which could compromise the performance of off-the-shelf priors and degrade the overall image quality in the long term. 

Since our approach incrementally increases the number of views through the generation of pseudo-ground-truth images, the computational time associated with 3D model training also escalates along the camera schedule. 
The current run time for the reconstruction of a single object takes around seven hours using a single 3090 RTX GPU. Nevertheless, our framework has a modular structure and it would be easy to replace the models used in each step. Especially, we may replace VolSDF~\cite{VolumeSDF} with more advanced methods~\cite{permutosdf, neus2} to accelerate the reconstruction process. Furthermore, we may adopt LoRA~\cite{lora} for accelerating DreamBooth~\cite{dreambooth}.

To address these issues, our future research will focus on exploring methods to minimize any artifacts and further refine the reconstruction process while improving the reconstruction time.

\begin{acks}
This research was supported by \grantsponsor{IITP}{IITP}{} grants funded by the Korea government (MSIT) (\grantnum[]{IITP}{IITP-2021-0-02068}, \grantnum[]{IITP}{IITP-2019-0-01906}), and the Starting growth Technological R\&D Program (TIPS Program, (No. \grantnum[]{TIPS}{S3200141})) funded by the \grantsponsor{MSS}{Ministry of SMEs and Startups (MSS, Korea)}{} in 2021.
\end{acks}

\bibliographystyle{ACM-Reference-Format}
\bibliography{bibliography}


\end{document}